\documentclass[times, review, 10pt]{elsarticle}
\usepackage{natbib}
\usepackage{subcaption}
\usepackage{algorithmic}
\usepackage{hyperref}

\usepackage{amsmath}
\usepackage{amssymb}
\usepackage{mathtools}
\usepackage{amsthm}
\usepackage{gensymb}
\usepackage{breqn}
\usepackage{gensymb}
\theoremstyle{plain}
\newtheorem{theorem}{Theorem}[section]

\theoremstyle{definition}

\theoremstyle{remark}

\usepackage[ruled,vlined]{algorithm2e}
\SetKwProg{Init}{initialize}{}{}
\usepackage{setspace}
\usepackage{xcolor}
\usepackage{nomencl}
\makenomenclature

\journal{Pattern Recognition}

\begin{document}

\begin{frontmatter}

%% Title, authors and addresses

%% use the tnoteref command within \title for footnotes;
%% use the tnotetext command for theassociated footnote;
%% use the fnref command within \author or \affiliation for footnotes;
%% use the fntext command for theassociated footnote;
%% use the corref command within \author for corresponding author footnotes;
%% use the cortext command for theassociated footnote;
%% use the ead command for the email address,
%% and the form \ead[url] for the home page:
%% \title{Title\tnoteref{label1}}
%% \tnotetext[label1]{}
%% \author{Name\corref{cor1}\fnref{label2}}
%% \ead{email address}
%% \ead[url]{home page}
%% \fntext[label2]{}
%% \cortext[cor1]{}
%% \affiliation{organization={},
%%            addressline={}, 
%%            city={},
%%            postcode={}, 
%%            state={},
%%            country={}}
%% \fntext[label3]{}

\title{Uncertainty-aware Efficient Subgraph Isomorphism using Graph Topology}

%% use optional labels to link authors explicitly to addresses:
%% \author[label1,label2]{}
%% \affiliation[label1]{organization={},
%%             addressline={},
%%             city={},
%%             postcode={},
%%             state={},
%%             country={}}
%%
%% \affiliation[label2]{organization={},
%%             addressline={},
%%             city={},
%%             postcode={},
%%             state={},
%%             country={}}

\author{Arpan Kusari and Wenbo Sun}

\affiliation{organization={University of Michigan Transportation Research Institute},%Department and Organization
            addressline={2901 Baxter Road}, 
            city={Ann Arbor},
            postcode={48109}, 
            state={Michigan},
            country={United States}}

\begin{abstract}
% complexity and our case
Subgraph isomorphism, also known as subgraph matching, is typically regarded as an NP-complete problem. This complexity is further compounded in practical applications where edge weights are real-valued and may be affected by measurement noise and potential missing data. Such graph matching routinely arises in applications such as image matching and map matching. Most subgraph matching methods fail to perform node-to-node matching under presence of such corruptions. 
% our proposal
We propose a method for identifying the node correspondence between a subgraph and a full graph in the inexact case without node labels in two steps - (a) extract the minimal unique topology preserving subset from the subgraph and find its feasible matching in the full graph, and (b) implement a consensus-based algorithm to expand the matched node set by pairing unique paths based on boundary commutativity. 
% results
To demonstrate the effectiveness of the proposed method, a simulation is performed on the Erdos-Renyi random graphs and two case studies are performed on the image-based affine covariant features dataset and KITTI stereo dataset respectively.
% advantages of our method
Going beyond the existing subgraph matching approaches, the proposed method is shown to have realistically sub-linear computational efficiency, robustness to random measurement noise, and good statistical properties. Our method is also readily applicable to the exact matching case without loss of generality. 
\end{abstract}   

%%Graphical abstract
\begin{graphicalabstract}
\includegraphics[width=\textwidth]{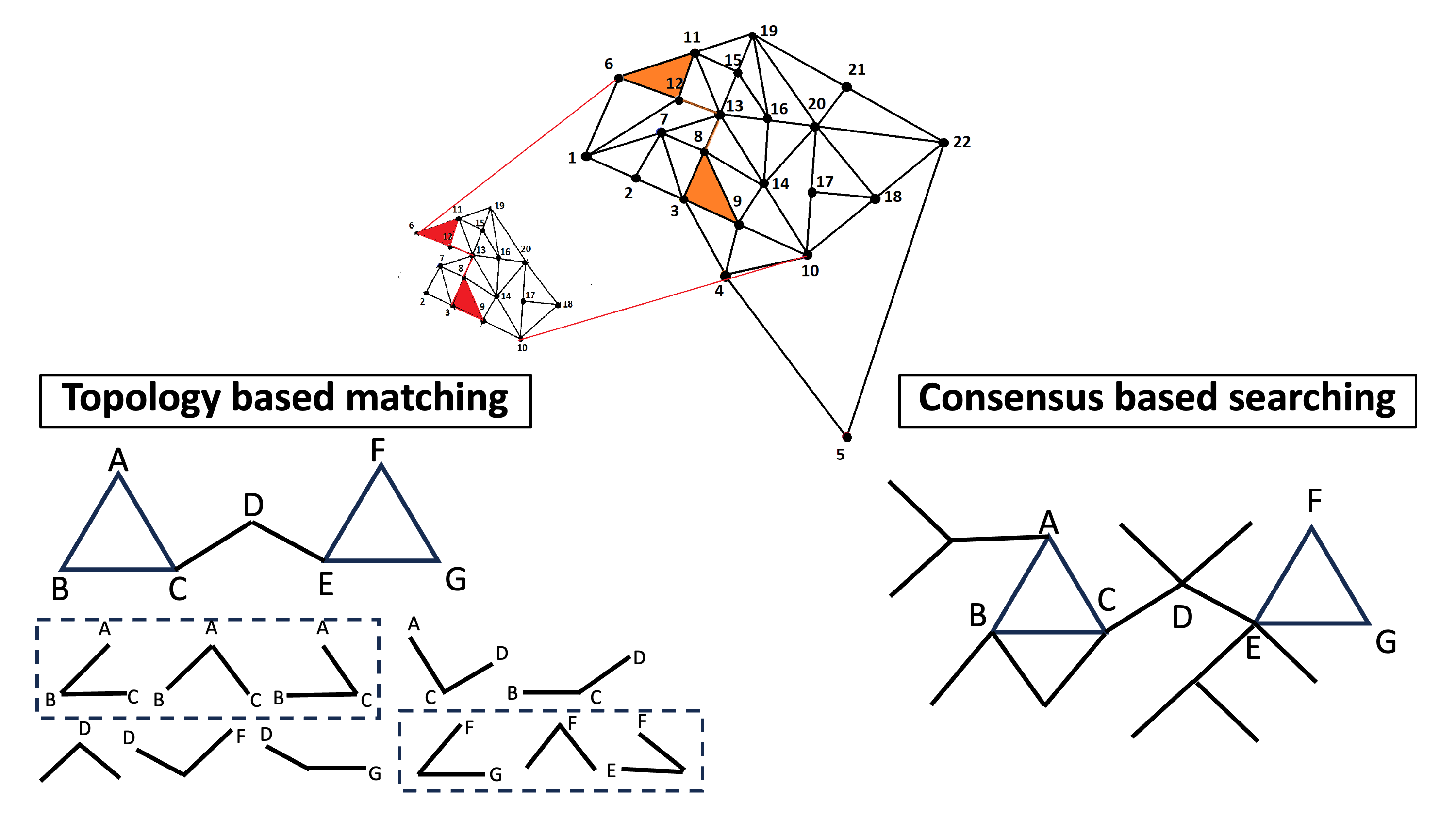}
\end{graphicalabstract}

%%Research highlights
\begin{highlights}
\item Matching a subgraph to a graph with measurement noise and randomly missing nodes arises in many real-world applications, like template matching and 3D registration. We introduce an analytical method that robustly matches graphs and subgraphs amidst uncertainties using topological constraints.
\item We employ a two-stage approach whereby initially, we match a minimum topological unit between a subgraph and a full graph and in the second step, we perform consensus-based searching.
\item We employ massively parallelizable pairwise edge matching operations along with topological constraints to provide fast and robust solutions in both steps.
\item We provide theoretical justifications for the steps of the algorithm and computational complexity measure. 
\item We provide real-life examples of template matching and stereo-matching along with simulation study to show the efficacy of the approach.
\end{highlights}

\begin{keyword}
%% keywords here, in the form: keyword \sep keyword

%% PACS codes here, in the form: \PACS code \sep code

%% MSC codes here, in the form: \MSC code \sep code
%% or \MSC[2008] code \sep code (2000 is the default)
topology, sub-graph matching, consensus
\end{keyword}

\end{frontmatter}

%% \linenumbers

%% main text
\section{Introduction}
\label{sec:intro}

% What is subgraph matching?
In recent years, the number of applications employing graphs have increased exponentially, with the scientific community using graphs to denote relationships between the features of interest \citep{goyal2018graph}. A fundamental problem arising in these applications is the subgraph isomorphism problem, also known as subgraph matching, whereby an optimal correspondence is sought between nodes of two given graphs, where one graph is derived from the other. We note here that graph matching is a specialized case of the subgraph matching problem when two graphs are of the same size. Optimality differs according to problem description and domain, but generally refers to the alignment of global graph structures, along with other node features when available. 

% previous references and differences between exact and inexact matching
Ullman in his seminal paper \citep{ullmann1976algorithm} showed that isomorphism can always be estimated using brute-force enumeration as a default strategy which amounts to a depth-first search. The backtracking strategy devised in the paper is used to significantly reduce the size of the search space. \cite{cordella2004sub} proposed an improvement of Ullman's algorithm, the VF2 algorithm, by utilizing a data structure during exploration to significantly reduce memory requirements along with five feasibility rules for pruning the search tree. \cite{eppstein2002subgraph} partitioned the planar graph into pieces of small tree-width and applied dynamic programming within each piece to solve the subgraph isomorphism problem in linear time. 
\cite{gouda2022scaling} provided an update on the Ullman algorithm using a novel sorting method for querying the vertices and constrain to L-levels of vertex neighborhoods to increase the speed while \cite{ma2024tps} provided a theoretical method for guiding the selection of vertex-searching orders, and solving the exact matching problem. Recently, \cite{ye2024efficient} proposed an efficient graph neural network-based path embedding framework to avoid false dismissals. All the above approaches solve the exact matching problem where the corresponding nodes and edges between the two graphs share same weights. Conversely, in many practical applications, the nodes can be corrupted by measurement noise and edges can be absent which renders exact matching ineffective.

To overcome the challenges due to measurement noise, inexact graph matching methods have been explored \citep{yan2016short}. In this case, the practitioners rely not on the exact correspondence but rather some similarity measures. The inexact graph matching problems can be formulated as a quadratic assignment problem (QAP) with varying types of affinity matrices. Lawler's QAP \citep{lawler1963quadratic} maximized the affinity score based on a second-order affinity. Matrix factorization techniques were then explored to provide taxonomy for graph matching \citep{zhou2015factorized}. Besides QAP, Graph edit distance was introduced to quantify the similarity between pairwise graphs \citep{sanfeliu1983distance, serratosa2014fast, serratosa2015speeding}. Different strategies have been proposed to solve the graph matching problems under the above formulations. The first set of inexact matching algorithms leverage learning algorithms to search for the optimal affinity matrix, through supervised \citep{caetano2009learning}, unsupervised \citep{leordeanu2012unsupervised}, or semi-supervised \citep{leordeanu2011semi} learning methods. The affinity matrix can be also estimated through a series of editing operations such as node and edge insertion, deletion or substitution, in order to attain an exact matching \citep{cordella1998subgraph, tsai1983subgraph, shapiro1985metric}. As is understood, in the case of subgraphs having a much lesser nodes than the full graphs, these methods are inefficient in solving the graph matching problems.

% inexact matching using optimization
The second set of inexact matching algorithms aimed to solve the NP-hard optimization problem through relaxations or approximations. \citep{egozi2012probabilistic} provides a probabilistic approach to spectral matching which provides a computationally efficient approach towards graph matching. \citep{suh2015subgraph} augments the integer quadratic problem (IQP) objective with a compactness prior to reduce the number of outliers matched. They utilize a Markov chain Monte Carlo (MCMC) to solve the optimization problem. \citep{adamczewski2015discrete} proposed a tailored Tabu search
for graph matching. \citep{yan2015discrete} developed a full discrete method for hyper-graph matching. \citep{aflalo2015convex, lyzinski2015graph} suggested to transform a graph matching problem to convex quadratic programming that enable different solvers. However, these approaches require construction of a weight matrix between the vertices of the two graphs and are thus, constrained by small graph sizes and pairwise interactions.

% topology preserving 
The third set of inexact matching algorithms set constraints on the feasible space of matching policies to enable efficient searching. The constraints were often imposed through geometry or feature compatibility. For example, \citep{shi2021robin} developed a theory of invariance to check if a subset of measurements were mutually compatible without explicitly solving the corresponding estimation problem. \citep{zaslavskiy2008path} proposed an approximate method as a quadratic assignment problem over the set of permutation matrices, where a permutation matrix is a binary matrix describing whether the particular nodes are matched or not. 
\cite{dahm2012topological} proposed topological features that can be embedded to nodes such as degree of nodes, distance to self, number of n-walks passing etc. along with iterative node elimination to speed up matching. \cite{zhang2022twig} provided a different topological structure of twig patterns and converted the graphs into twig patterns to provide the matching. \cite{liu2019g} provided an inexact matching technique based on constructing a lookup-table based on the graph connectivity information to make the finding faster including missing nodes and edges, but do not consider the noise in weights along with the cost of constructing such a lookup-table. There have been other approaches involving constructing permutation matrices under specific real-valued approximations \citep{cho2010reweighted, zhou2012factorized} but they have problems where there are poor geometric alignments with high deformations.  \citep{byrnetopology} proposed a graph matching technique which preserves the global topological structure by finding an optimal simplicity chain map. Although the author provides an explanation of how this technique can be extended for higher topologies, the formulation shows a matching between the nodes and the edges between the nodes to constrain the homology. For these approaches, even though the topology is constrained as a result of the optimization, the topology constraint is not inherently employed in the matching.

% our contributions
In this work, we specifically focus on graph matching problems with considering measurement noise on edge weights without the presence of node labels, which are demanded in a variety of applications \citep{bunke2000graph} such as character recognition \citep{lu1991hierarchical, rocha1994shape}, shape analysis \citep{cantoni19982, lourens1998biologically}, chemical structure analysis \citep{balaban1985applications}, etc. Graph topology provides a description of graph connectivity as the basic structure of a graph. Our intuition is that matching the unique minimal unit that preserves a pre-specified graph topology between the subgraph and full graph improves the matching efficiency and robustness. Post to the topology-based matching, the remaining unmatched edges in the subgraph are further matched to the corresponding edges in the full graph. However, there are three fundamental questions that arise - (a) what is the minimal topology-preserving unit? (b) how do we guarantee the matching uniqueness? (c) how do we match the remaining edges after the initial match has occurred? 

The idea of matching topology between subgraph and full graph is inspired from the random sample consensus (RANSAC) literature \citep{fischler1981random, derpanis2010overview}. 
RANSAC and its variants provide a method for robustly estimating parameters of a model based on a set of observed data with significant outliers. The algorithm works in two steps - generating a hypothesis with the minimum number of data points and checking whether the rest of the points follow this hypothesis known as consensus set. However, there are important differences between RANSAC and our proposed approach - our initial matching employs a hierarchical constraint setup based on topology, composed of specific cyclic and connectivity constraints. Providing these specific constraints helps in yielding unique candidates even in the presence of significant outliers. Also these constraints are independent of each other and therefore, can be run in parallel. This significantly reduces the suitable candidates and takes away a lot of the computational burden. 

We would also like to mention that while we are aware of the various efforts in deep learning based matching, our paper is not utilizing a learning-based solution and our goal is to provide a robust analytical solution to the graph matching problem. While deep learning has obviously shown superior performance to traditional statistical methods, there is a requirement of large amount of labeled training data which might not always be available. Since, we focus on utilizing topology of constrained features in the graphs to determine correspondences, we can provide an uncertainty-aware solution even when matching a single subgraph to a full graph.

The primary contributions of this work are: 
\begin{itemize}
    %\item \textbf{Inexact matching in absence of node labels:}
    %Almost all of the subgraph matching methods rely on node labels to provide some information regarding the nodes. In absence of such information, these methods fail to find feasible subgraph matches. Our method therefore, fills this gap and provides a matching method which does not need to incorporate labels for matching. 
    \item \textbf{Globally consistent matching in the presence of large amounts of outliers}: Almost all subgraph matching algorithms rely on a single optimization step to search for the node correspondence minimizing the matching loss, which may lead to invalid matching since the true matching may not result in the minimal matching loss under measurement noises. Our proposed approach provides robust global matching under large noise by defining a graph-topology-based structure in the subgraph and searching for feasible matches of nodes sharing the same graph-topology-based structure between the full graph and subgraph. %Given the feasible initial match, the consensus based searching expands the search for new matched nodes in a breadth-first tree approach to collect the number of nodes which can be matched uniquely. %This formulation takes into account the noise in the edges by statistically modeling them and proving theoretically that this can converge to the true matching. 
    \item \textbf{Providing efficiency in subgraph matching}:  We strive for efficiency in our matching solution by precomputing set of edge pairs for both subgraph and full graph and then selecting `top-k' matches in the full graph for each edge pair in the subgraph. In the topological matching step, we utilize topological constraints to drive the solution towards finding an unique set of matching nodes. In the consensus searching step, we further select matches based on path-wise matching until an unique match is established.  This particular approach leads to a fast convergence whereby a wrong initial matching leads to a weaker final match set. Parallelizing the two steps can yield further efficiencies without sacrificing accuracy.
\end{itemize}

The rest of the paper is organized as follows: Section \ref{sec:method} presents the mathematical formulation of the graph matching problem and our proposed approach. Section \ref{sec:implementation} provides the implementation details of the proposed approach. Section \ref{sec:simulation} provides a robust simulation study to compare the strengths of our algorithm and in sections \ref{sec:case} and \ref{sec:stereo}, we evaluate performance on homology preserving subgraph matching for image registration applications. Section \ref{sec:conc} provides the conclusions.

\section{Method}\label{sec:method}

\begin{tabular}{|c |l|}
    \hline
    \multicolumn{2}{|c|}{Nomenclature} \\
    \hline
    $\alpha$ &  Type-I error for matching\\
    $\bar{c}$ & Average number of nodes among topology-preserving units \\
    $\bar{d}_f$ & Average degree in $G_f$\\
    $\delta_{s,c}$ & Topology-preserving unit \\
    $\epsilon$ & Observation noise \\
    $\eta$ & Intermediate set for consensus-based searching \\
    $\mathcal{T}$ & Feasible matching set\\
    $\phi$ & Matching policy \\
    $\tau$ & Matching loss threshold \\
    $d$ & Distance measure between edge weights \\
    $G=(V,E,w)$ & Graph with sets of nodes, edges and weights \\
    $ G_f, G_s$ & Faull graph and subgraph \\
    $m_f$ & Number of p simplices in $G_f$ \\
    $Q$ & Overall subgraph matching loss \\
    $ X, X_v$ & Matching matrix and its vectorization\\
    \hline
\end{tabular}

% \nomenclature{\(G=(V,E,w)\)}{Graph with sets of nodes, edges and weights}
% \nomenclature{\(G_f, G_s\)}{Full graph and subgraph}
% \nomenclature{\(\phi\)}{Matching policy}
% \nomenclature{\(\epsilon\)}{Observation noise}
% \nomenclature{\(d\)}{Distance measure between edge weights}
% \nomenclature{\(Q\)}{Overall subgraph matching loss}
% \nomenclature{\(X, X_v\)}{Matching matrix and its vectorization}
% \nomenclature{\(\tau\)}{Matching loss threshold}
% \nomenclature{\(\mathcal{T}\)}{Feasible matching set}
% \nomenclature{\(\delta_{s,c}\)}{Topology-preserving unit}
% \nomenclature{\(\eta\)}{Intermediate set for consensus-based searching}
% \nomenclature{\(\alpha\)}{Type-I error for matching}
% \nomenclature{\(\bar{c}\)}{Average number of nodes among topology-preserving units}
% \nomenclature{\(m_f\)}{Number of p simplices in $G_f$}
% \nomenclature{\(\bar{d}_f\)}{Average degree in $G_f$}
% \printnomenclature

\subsection{Problem formulation and preliminaries}
We start with a formal definition of graphs. A graph is denoted by $G=\left(V,E,w\right)$, where $V$, $E$ and $w$ represent the set of nodes, the set of edges and edge weights respectively. Let $|\cdot|$ denote the number of elements in a set. Each node is denoted by a node index $v\in\left\{1,...,|V|\right\}$. Each edge is expressed as $e=(i,j)$ indicating that nodes $i$ and $j$ are connected by an edge. The edge weight $w:(i,j)\rightarrow \mathbb{R}$ is a function mapping an edge to the real-valued weight. We discuss more about the choice of weights in Sec. \ref{subsec:weights}.

In our problem setting, let $G_f=\left(V_f,E_f,w\right)$ and $G_s=\left(V_s,E_s,w\right)$ denote the full graph and subgraph of interest, respectively. 
%Here the subscript ``f'' represents ``full'' while the subscript ``s'' represents ``sub''.
The objective of the subgraph matching problem is to seek a matching policy $\phi$, that projects each node $v\in V_s$ to the corresponding node $\phi(v)\in V_f$. In some graph matching literature, the policy function can be also expressed as a $|V_f|\times|V_s|$ permutation matrix $X$, whose $(i,j)$-th element equals to $1$ if $\phi(j)=i$. Without ambiguity, for any $e=(i,j)\in E_s$, we define $\phi(e)=\left(\phi(i),\phi(j)\right)\in E_f$ as the matched edge of $e$. In this paper, we consider the situations where weights in the full graph are the ground truth, and the weights in the subgraph are noisy observations from the full graph, that is to assume:
\begin{equation}
    w(e_s)=w\left(\phi(e_s)\right)+\epsilon,
    \label{eq:normal}
\end{equation}
where $\epsilon$ is the random noise following normal distribution $N(0,\sigma^2)$ with unknown variance $\sigma^2$. We assume $\epsilon$'s are independent among different pairs of matched edges. 

Subgraph matching can be formulated as an optimization problem. Under a pre-specified distance measure $d(\cdot,\cdot)$ of edges, the overall matching loss can be quantified as a $|V_f||V_s|\times |V_f||V_s|$ matrix $Q$ whose $(i+|V_s|j,i'+|V_s|j')$-th element equals to $d\left((i,i'),(j,j')\right)$. Let $X_v$ denote the vectorization of $X$ by stacking its columns on top of one another. Let $X_i$ and $X_{(j)}$ denote the $i$-th row and $j$-th column vectors of $X$, respectively. Let $\mathbb{I}$ be the vector of ones. The objective function is written as:
\begin{equation}
    \widehat{X}_v=arg\,\min_{X_v} X_v^T Q X_v
    \label{eq:quadratic}
\end{equation}
subject to
\begin{eqnarray}
    \mathbb{I}^T X_i&=&1\nonumber\\
    X_{(j)}^T\mathbb{I}&=& 1,\nonumber
\end{eqnarray}

The exact solution to the optimization problem in Eq.(\ref{eq:quadratic}) can be obtained through an NP-complete non-convex integer programming. The problem can be solved via a searching algorithm (e.g. the Hungarian algorithm) or approximately solved via an integer programming solver (e.g. interior point optimizer). There are two main limitations of these existing approaches. (i) They require a computational time of $O(|V_f|^3)$, which often exceeds the computation budget for large-scale graphs. (ii) The existing methods are designed based on the loss of edge-to-edge matching. Simply minimizing the overall matching loss in Eq.(\ref{eq:quadratic}) may lead to erroneous results due to the measurement noise. 

To overcome the above limitations, we propose a consensus-based approach that performs graph matching on a set of nodes with topological constraints, which introduces few key innovative components. First, topological constraints are employed for matching a set of nodes. Compared to the edge weights, the connectivity between nodes imposed by the topological constraints are invariant to the measurement noise with probability one (in the case a positive-valued edge weight is disturbed to zero). Moreover, simultaneously matching multiple nodes which satisfy a topological constraint can reduce computational time while shrinking the probability of mismatching due to the measurement noise. Second, instead of searching for the unique solution to Eq.(\ref{eq:quadratic}) in a single step, we would like to bind the feasible matching policies in a feasible set based on the revised version of Eq.(\ref{eq:quadratic}) subject to some topological constraints and then shrink the feasible set with accumulating more nodes to the topology structure until a unique matching policy is reached. In this way, the proposed method will be able to cover the true matching policy in the feasible region with a high probability and identify it as the unique solution when a sufficient number of nodes are introduced to the topology structure. It is required to define the topological constraint and feasible set to complete the proposed method. While we do not formally show the proof, the exact matching case can be easily shown to be a special case of our inexact matching algorithm. We will elaborate the technical details in the following subsections. 

\subsection{Topology-based matching}
\begin{figure}[h]
  \centering
  \includegraphics[height=3in]{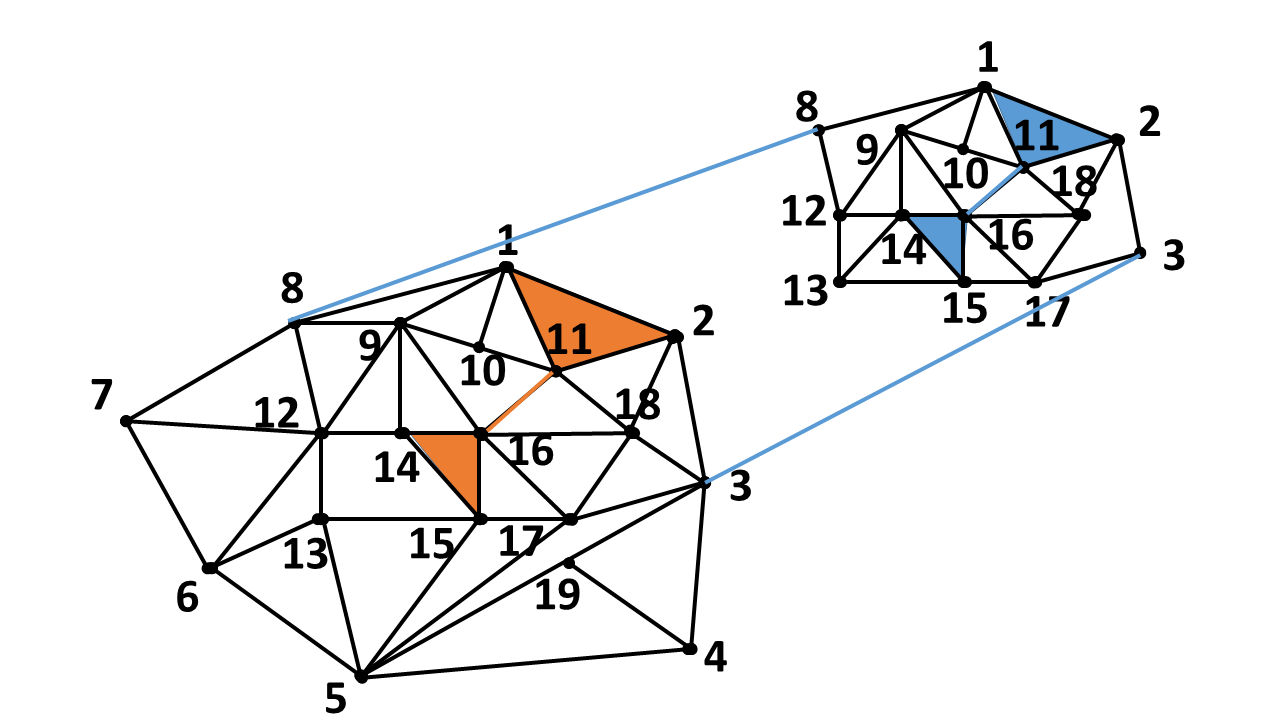}
  \caption{Illustration of topological matching unit comprised of two simplexes connected by the shortest path in initial matching. Left panel: full graph $G_f$ with the unique matching simplexes highlighted in orange. Right panel: subgraph $G_s$ with two $2$-simplexes and the path highlighted in blue. Certain edges are absent both in subgraph and full graph.}
  \label{fig:illu}
\end{figure}

In the first step, we aim to match a subset of $V_s$ to $V_f$, which is defined as a topology-preserving unit. Different from the node-to-node matching in Eq.(\ref{eq:quadratic}), the graph matching is conducted based on multiple mutually exclusive subsets of $V_s$ following a topology concept - p-simplex, which is defined as the convex hull of $p+1$ affinely independent nodes. For example, a $2$-simplex represents a triangle, and a $3$-simplex represents a tetrahedron in a graph. The topology-preserving unit in the initial matching is defined as a pair of connected p-simplexes in the graph along with the connecting path. We choose p-simplexes as topology-preserving units due to the computational load incurred in the first step. Specifically, an overly low occurrence of the topology unit in the subgraph can result in few nodes matched in the initial step while an overly high occurrence of the topological unit may increase the searching complexity. To guarantee the validity of the topology-preserving unit, two basic assumptions are made while formulating our approach. 
\begin{itemize}
    \item $G_s$ is a connected graph i.e. there do not exist two or more separate clusters of nodes without edges in between, and
    \item there exists more than two p-simplexes in $G_s$.
\end{itemize}
An illustrative example of a topology-preserving unit is displayed in Figure~\ref{fig:illu}, where the two 2-simplexes in the subgraph on the left panel (nodes $\{3, 8, 9\}$ and nodes $\{6, 11, 12\}$) and the shortest path in between (nodes $\{8, 13, 12\}$) are considered as the topology-preserving unit and matched with the corresponding nodes in the full graph. Let $\Delta_p$ denote the set of $p$-simplex. Let $\Theta$ denote the topology-preserving unit, the next step is to seek a matching policy $\phi$ or equivalently a vectorized matching matrix $X_v$ giving the $\Theta$ matching nodes in $G_f$.

Here we define a general feasible set of matching policies for any given subset of $V$, which can be naturally applied to $\Theta$. Let $V$ denote any subset of $V_s$. Let $Q(V)$ denote the sub-matrix of $Q$ that collects the distance measures between $V$ and $V_f$. The feasible set of vectorized matching matrix $\widetilde{X}_v(V)$ for the node set $V$ is defined by revising Eq.(\ref{eq:quadratic}) as:
\begin{equation}
    \widetilde{X}_v(V)=\left\{X_{v}(V)\big| X_v(V)^T Q(V) X_v(V) \leq \tau_c\right\},
    \label{eq:feasible1}
\end{equation}
where $\tau_c$ is a pre-specified threshold determined by the set size $c$ of $V$. Revising the optimization problem into a feasible set searching problem enhances the robustness of the matching under measurement noise. Under the statistical property developed in the later subsection, the number of elements in $\widetilde{X}_v(V)$ will shrink as the set size $c$ increases. For a $c$ large enough, a unique element will be eventually identified.

Now we substitute the notation $V$ by $\Theta$ that incorporates topology structure and search for its initial matching in $G_f$. We find the initial match by finding simplexes in the full graph with similar edge weights to the simplexes in the topological unit, $w(\theta_f) - w(\theta_s) \simeq 0$ for each simplex $\theta$ and then matching the connecting path $p_f$ between the simplexes in the full graph with the subgraph $p_s$ such that $w(p_f) - w(p_s) \simeq 0$. In case of a possible path not being present, the match is rejected. % However, it is easy to see that the number of combinations in this approach can be large, especially with the length of the connecting path. 

To this end, we denote the set of vectorized matching matrices satisfying the above four constraints by $\widehat{\mathcal{T}(\Theta)}$. The feasible set is formally defined as:
\begin{equation}
    \widetilde{X}_v(\Theta)=\left\{X_{v}(\Theta)\in \widehat{\mathcal{T}(\Theta)}\big| X_{v}(\Theta)^T Q(\Theta) X_{v}(\Theta) \leq \tau_c\right\}.
    \label{eq:feasible2}
\end{equation}

Note that the distance measure $Q(\Theta)$ is invariant to possible rotation and scaling in real graphs. 

The first step is implemented via a depth-first searching of feasible elements based on Eq.(\ref{eq:feasible2}), as described in Algorithm \ref{a:topology}. The topology structure and shortest path information are used to prune the search. In particular, the following searching algorithm is designed to find all feasible matching policies for the given topology-preserving units, which will serve as a foundation for the consensus-based searching elaborated in the next subsection.

\begin{algorithm}[tb]
    \SetAlgoLined
    \SetKwInOut{Input}{input}
    \SetKwInOut{Output}{output}
    
    \Input{The subgraph $G_s$ and the full graph $G_f$, initial parameters $\alpha$ and $\sigma$, number of tries $n$}
    \Init{$\Theta=\left\{\right\}$,$\widetilde{X}_v(\Theta)=\left\{\right\}$}

    \For{$i \in 1\dots n$}{
        Randomly draw a pair of p-simplexes $\delta_1 \neq \delta_2$ from $G_s$\\
        Calculate the shortest path $p(\delta_1,\delta_2)$ between $\delta_1$ and $\delta_2$\\
        $\theta \gets \delta_1 \cup p(\delta_1,\delta_2) \cup \delta_2$\\
        $\tilde{\delta}_{1,f}\gets$ all feasible matches to $\delta_1$ from $G_f$ based on Eq.(\ref{eq:feasible2})\\
        $\tilde{\delta}_{2,f}\gets$ all feasible matches to $\delta_2$ from $G_f$ based on Eq.(\ref{eq:feasible2})\\
        \For{$(\delta_{1,f},\delta_{2,f})\in \tilde{\delta}_{1,f} \times \tilde{\delta}_{2,f}$}
        {
            \For{Path $p(\delta_{1,f},\delta_{2,f})$ between $\delta_{1,f}$ and $\delta_{2,f}$ having the same length as $p(\delta_1,\delta_2)$}{
                $v=\delta_{1,f}\cup p(\delta_{1,f},\delta_{2,f}) \cup \delta_{2,f}$\\
                \If{$d(v,\theta)\leq \tau_c$} {
                    $\Theta\gets \Theta\cup  \theta$\\
                    $\phi(\theta) \gets v$\\
                    Calculate $X_{v}(\Theta)$ based on $\phi(\Theta)$\\
                    \For{$X_v(\theta) \in \widetilde{X}_v(\theta)$} {
                        $X_v(\Theta) \gets X_v(\Theta) \cup X_{v}(\theta)$\\
                    }
                }
            }
        }
    }
    \Output{Topology-preserving unit $\Theta$ and the feasible matching $\widetilde{X}_v(\Theta)$}
    \caption{Topology-based matching}
    \label{a:topology}
\end{algorithm}

\subsection{Consensus-based searching}
After the initial matching of the topology preserving nodes in $\delta_{s,c}$, a consensus-based algorithm is developed to sequentially match the remaining nodes in $V_s\backslash \delta_{s,c}$. The basic concept is to seek for the feasible matching of paths whose starting nodes are in $\delta_{s,c}$ and the rest nodes are in $V_s\backslash \delta_{s,c}$. Instead of matching the topology-preserving unit in the previous step, we aim to match paths of length $q$ based on the feasible set defined in Eq.(\ref{eq:feasible2}). The feasible set can be narrowed down as the path length $q$ grows large. We start with an empty node set $\eta_{s,v}$ with $v=0$ and expand the set by searching for feasible matching of paths in $V_s\backslash \delta_{s,c}$ until it contains a unique element. The algorithm is terminated when no new nodes can be matched between $G_s$ and $G_f$. Specifically, we propose the following consensus-based searching algorithm, Algorithm \ref{a:consensus}. 

\begin{algorithm}[tb]
    \SetAlgoLined
    \SetKwInOut{Input}{input}
    \SetKwInOut{Output}{output}
    
    \Input{Topology-preserving unit $\delta_{s,c}$, unmatched node set $V_s\backslash \delta_{s,c}$, initial parameters $\alpha$ and $\sigma$}
    
    \Init{$v=0$, $\eta_{s,v}=\left\{\right\}$, $\widetilde{X_v}(\eta_{s,v})=\left\{\right\}$}

    \While{$\eta_{s,v} \neq V_s\backslash \delta_{s,c}$}{
        Sample an edge $e_s\in E_s$ incidence to a node in $\eta_{s,v}$ and a node $a_s\in V_s\backslash \eta_{s,v}$\\
        $\eta_{s,v} \gets \eta_{s,v} \cup \{v:v\in e_s\}$\\
        $v\gets v+1$\\
        $v'\gets 0$\\
        $\tilde\eta_{s,v'}=\left\{\right\}$\\
        \While{$\exists$ an edge $e_s'\in E_s$ incidence to $a_s$ and a node $b_s\in V_s\backslash \eta_{s,v}$}{
            $\tilde\eta_{s,v'} \gets \tilde\eta_{s,v'} \cup e_s'$\\
            $v'\gets v'+1$\\
            $a_s\gets b_s$\\
            Compute feasible matching for $\tilde\eta_{s,v'}$ in $V_f$ according to Eq.(\ref{eq:feasible2}), store it in $\widetilde{X_v'}(\eta_{s,v'})$\\
            \If{$\widetilde{X_v'}(\eta_{s,v'})$ is unique}{
                $\eta_{s,v}\gets \eta_{s,v}\cup \tilde\eta_{s,v'}$\\
                $v\gets v+v'$\\
                $\widetilde{X_v}(\eta_{s,v})\gets \widetilde{X_v}(\eta_{s,v})\cup \widetilde{X_v'}(\eta_{s,v'})$\\
                break\\
            }
        }
    }
    \Output{Node correspondence for $\widetilde{X_v}(\eta_{s,v})$}
    \caption{Consensus-based searching}
    \label{a:consensus}
\end{algorithm}

\subsection{Theoretical Guarantees}\label{ss:theorem}
Theorems are developed to evaluate the false positive and false negative rates in Subsection~\ref{ss:theorem}. When all the edge weights are matched, the distance measure $X_{v}(\Theta)^T Q(\Theta) X_{v}(\Theta)$ is the arithmetic mean of $c$ independent normal random variables, which shrinks to zero as $c$ grows large according to the law of large numbers and thus guarantees the identification of the true matching policy. 
Under this statistical property, we propose to set the threshold $\tau_c$ as:
\begin{equation}
    \tau_c=\frac{\Phi^{-1}(1-\alpha/2)}{\sqrt{c}}\sigma,
    \label{eq:threshold}
\end{equation}
where $\Phi^{-1}(1-\alpha/2)$ represents the $1-\alpha/2$ quantile of the standard normal distribution. Here $1-\alpha$ is interpreted as the coverage rate of the true matching by the designed feasible set, and can be specified by the users based on the requirement on the coverage rate. %An estimate of $\sigma$ should be provided when it is not available to the user, which will be discussed in Section~\ref{sec:simulation}. 

The first theorem shows that the feasible sets in Eq.(\ref{eq:feasible2}) will cover the true matching nodes with a pre-specified probability $1-\alpha$, which guarantees the identification of the true matching nodes by applying the proposed method.
\begin{theorem}
    Let $X_{v}^*(\Theta)$ denote the true vectorized matching matrix corresponding for the node set $\Theta$. The feasible set covers the true matching nodes with probability $1-\alpha$, i.e.,
    % \vspace{-0.5cm}
    \begin{equation}
        \mathbb{P}\left[X_{v}^*(\Theta)\in \widetilde{X}_{v}(\Theta)\right]=1-\alpha.
    \end{equation}
    % \vspace{-2.0cm}
    \label{theorem:feasible_set}
\end{theorem}

On the other hand, we would like the proposed feasible set to exclude any irrelevant node set, especially when the number of nodes $c$ grows large. The theorem is formally established as follows.
\begin{theorem}
    Let 
    $$\mu=\inf_{(e_s,e_f)\in E_s\times E_f}\left|\mathbb{E}[w(e_s)]-\mathbb{E}[w(e_f)]\right|/\sigma>0,$$ where $e_s$ and $e_f$ are not matched. The probability of excluding an irrelevant vectorized matching matrix $X_v$ from $\widetilde{X}_{v}(\Theta)$
\begin{align}
     \mathbb{P}\left[X_v\notin \widetilde{X}_{v}(\Theta)\right]>\Phi\left(\frac{-\Phi^{-1}(1-\alpha/2)-\mu\sqrt{c}}{\sqrt{2c}}\right)+ \notag\\ 
       1-\Phi\left(\frac{\Phi^{-1}(1-\alpha/2)-\mu\sqrt{c}}{\sqrt{2c}}\right) \notag,
\end{align}
    \label{theorem:other_match}
\end{theorem}

\begin{align}
\mathbb{P}\left[\left|Z_{1,2}\right|>\tau_c\right] = \Phi\left(\frac{-\tau_c-\mu\sigma}{\sqrt{2}\sigma}\right)+1-\Phi\left(\frac{\tau_c-\mu\sigma}{\sqrt{2}\sigma}\right)\nonumber\\
    = \Phi\left(\frac{-\Phi^{-1}(1-\alpha/2)-\mu\sqrt{c}}{\sqrt{2c}}\right)+ \notag\\ 
    1-\Phi\left(\frac{\Phi^{-1}(1-\alpha/2)-\mu\sqrt{c}}{\sqrt{2c}}\right) .
\end{align}

Note that $\mathbb{P}\left[\left|Z_{1,2}\right|>\tau_c\right]$ is a monotonic increasing function with $c$, which implies that the feasible solution proposed can exclude irrelevant nodes when $c$ grows large.

\subsection{Computational complexity}
The computational complexity of the proposed method can be estimated as follows. Suppose there exist $k$ common $p$-simplexes between the subgraph and full graph. Let $\bar{c}$ denote the average number of nodes in each $\delta_{s,c}$, $m_f$ denote the total number of $p$ simplexes in $G_f$, and $\bar{d}_f$ denote the average degree in $G_f$. The computational cost in the initial matching step is $\mathcal{O}\left(m_f k\bar{d}_f\left(\bar{c}-2(p+1)\right)\right)$, where the multiplication of the first two terms evaluates the computation time for matching the $p$-simplexes while the last two terms quantifies the computation time for matching the shortest path in between. The computation time for the consensus step is dominated by the efforts in path-wise matching of the remaining nodes after the initial matching, which is $\mathcal{O}\left(\left(n_s-k\left(p+\bar{c}-1\right)\right)\bar{d}_f\right)$.

The total computation complexity of the proposed algorithm is expressed in
\begin{equation}
\mathcal{O}\left(m_f k\bar{d}_f\left(\bar{c}-2(p+1)\right) + \left(n_s-k\left(p+\bar{c}-1\right)\right)\bar{d}_f\right).
\label{eq:complexity}
\end{equation}
It is worth noting that Eq(\ref{eq:complexity}) is upper bounded by $\mathcal{O}\left((n_f+m_f) \bar{d}_f\right)=\mathcal{O}\left(n_f^p\right)$ in the worst case scenario where the subgraph and full graph have the same size and both of them are fully connected. In this case, the number of p-simplexes is $m_f=\mathcal{O}(n_f^p)$, which greatly increase the computational load. However, this case is not commonly seen in real applications where the graphs are relatively sparse. For a sparsely connected graph with a smaller $m_f$ and $\bar{d}_f$ and $n_s \ll n_f$, the computational time is sub-linear with respect to $n_f$.
\section{Implementation details}
\label{sec:implementation}
\subsection{Choice of weights}
\label{subsec:weights}
The proposed approach hinges on choosing a proper set of weights for the matching problem. Our goal is for these weights to be invariant under affine transformations. We draw inspiration from the field of computational chemistry and molecular modeling, where knowledge-based potentials serve as empirical energy functions to predict the behavior of protein conformations. In our approach, we use three different potentials based on the specific type of affine transformation we aim to estimate:
\begin{itemize}
    \item Distance-based potentials: These potentials are constructed based on the observed distribution of distances between the nodes, expressed as:
    \begin{equation}
        U(d) = -k_B T ln\Bigg(\dfrac{\mathbb{P}\left[d\right]}{\mathbb{P}_{ref}\left[d\right]}\Bigg)
    \end{equation}
    where $\mathbb{P}\left[d\right]$ is the observed probability of distance $d$, $\mathbb{P}_{ref}\left[d\right]$ is the reference probability, $k_B$ is the Boltzmann's constant and $T$ is the temperature. We utilize distance-based potentials for affine transformations such as rotations or translations. 
    \item Angle-based potentials: These capture the directional preferences between the nodes and can be given as:
    \begin{equation}
        U(cos(\phi)) = -k_B T ln\Bigg(\dfrac{\mathbb{P}\left[cos(\phi)\right]}{\mathbb{P}_{ref}\left[cos(\phi)\right]}\Bigg)
    \end{equation}
    where $\mathbb{P}\left[cos(\phi)\right]$ is the observed probability distribution of the cosine of the angle $\phi$ between two edges. This potential is obviously better suited to the edge pairwise weight combination. We utilize this potential for scaling operation.
    \item Combined distance and angle potentials: To provide a more robust weight choice, we also combine the distance and angle information in the form of $\mathbb{P}\left[|d_{e1} - d_{e2}|cos(\phi)\right]$ where $d_{e}$ represents the distance between nodes for the particular edge and $\phi$ is the angle between the edges.
\end{itemize}
In these experiments, we use a reference probability from a uniform distribution and specific Boltzmann constant to compute these potentials.

\subsection{Computing edge-pair shingles}
To improve the efficiency in calculating $\widetilde{X}_v(\Theta)$, we utilize a parallel matching process using edge pair shingles. We define the shingle as a connected edge pair as ${<e}_{i,j}, e_{j,k}{>}$ where $i$, $j$ and $k$ are the nodes respectively, inspired from Locally Sensitive Hashing \citep{indyk1998approximate}. For all shingles ${<}e_{i,j}, e_{j,k}{>} \in \Theta$, we choose the top-k matches in the full graph as:
\begin{equation}
    \mathcal{T}(\Theta) = \min_k~~|w(e_{m,n}) - w(e_{i,j})|~+~|w(e_{n,p}) - w(e_{j,k})|
\end{equation}
where $m$, $n$ and $p$ are connected nodes in the full graph. 
Given these matches, we impose the following four topology constraints on the matches to weed out the incorrect assignments from $\mathcal{T}(\Theta)$:
\begin{itemize}
    \item For any edge-pair shingle ${<}e_{i,j}, e_{j,k}{>}$ belonging to a $p$-simplex $\delta_p\in\Delta_p$, the matched shingle ${<}e_{m,n}, e_{n,p}{>} \in \mathcal{T}(\Theta)$ must belong to a p-simplex as well. 
    \item For two shingles ${<}e_{i,j}, e_{j,k}{>}$ and ${<}e_{j,i}, e_{i,k}{>}$ belonging to a p-simplex, the corresponding matched shingle pair in the full graph need to be from the same p-simplex in the full graph. These two constraints guarantee that a p-simplex in the subgraph will match with a p-simplex in the full graph.
    \item For any shingle containing edge $e_{i,j}$ in the subgraph, the matched shingle in the full graph should have a common edge.
    \item The longest simple path in $\Theta$ must be maintained between $G_s$ and $G_f$.  This constraint implies the matched nodes from any path realization of a p-simplex in the subgraph share the same node set.
\end{itemize}

\section{Simulation study}
\label{sec:simulation}

\subsection{Erdos-Renyi Graph matching}
\begin{figure*}[tp]
  \centering
    \begin{subfigure}[t]{0.45\textwidth}
        \centering
        \includegraphics[height=1.5in]{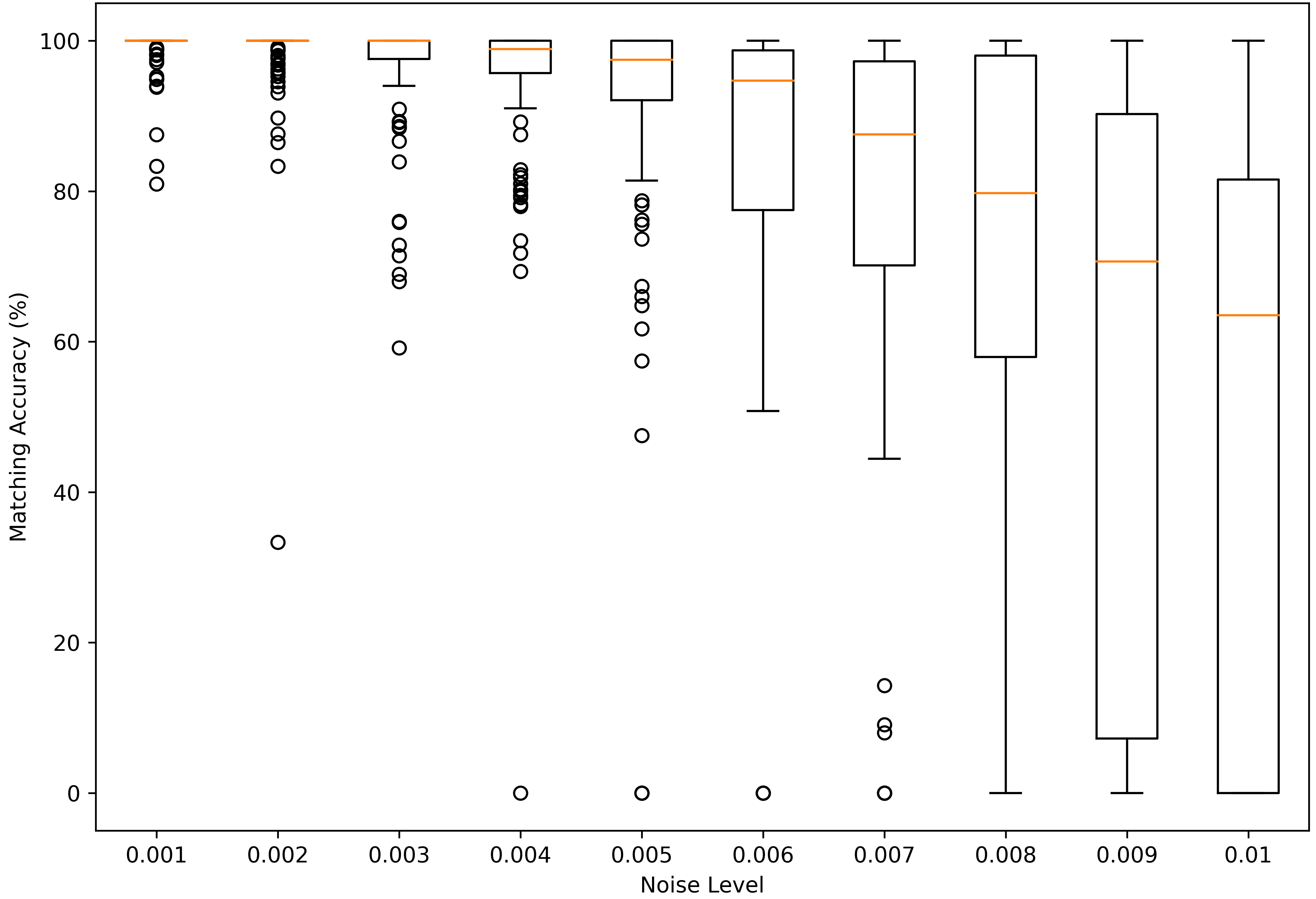}
        \caption{Box plot of the matching accuracy under different noise levels.}
    \end{subfigure}
    \begin{subfigure}[t]{0.45\textwidth}
        \centering
        \includegraphics[height=1.5in]{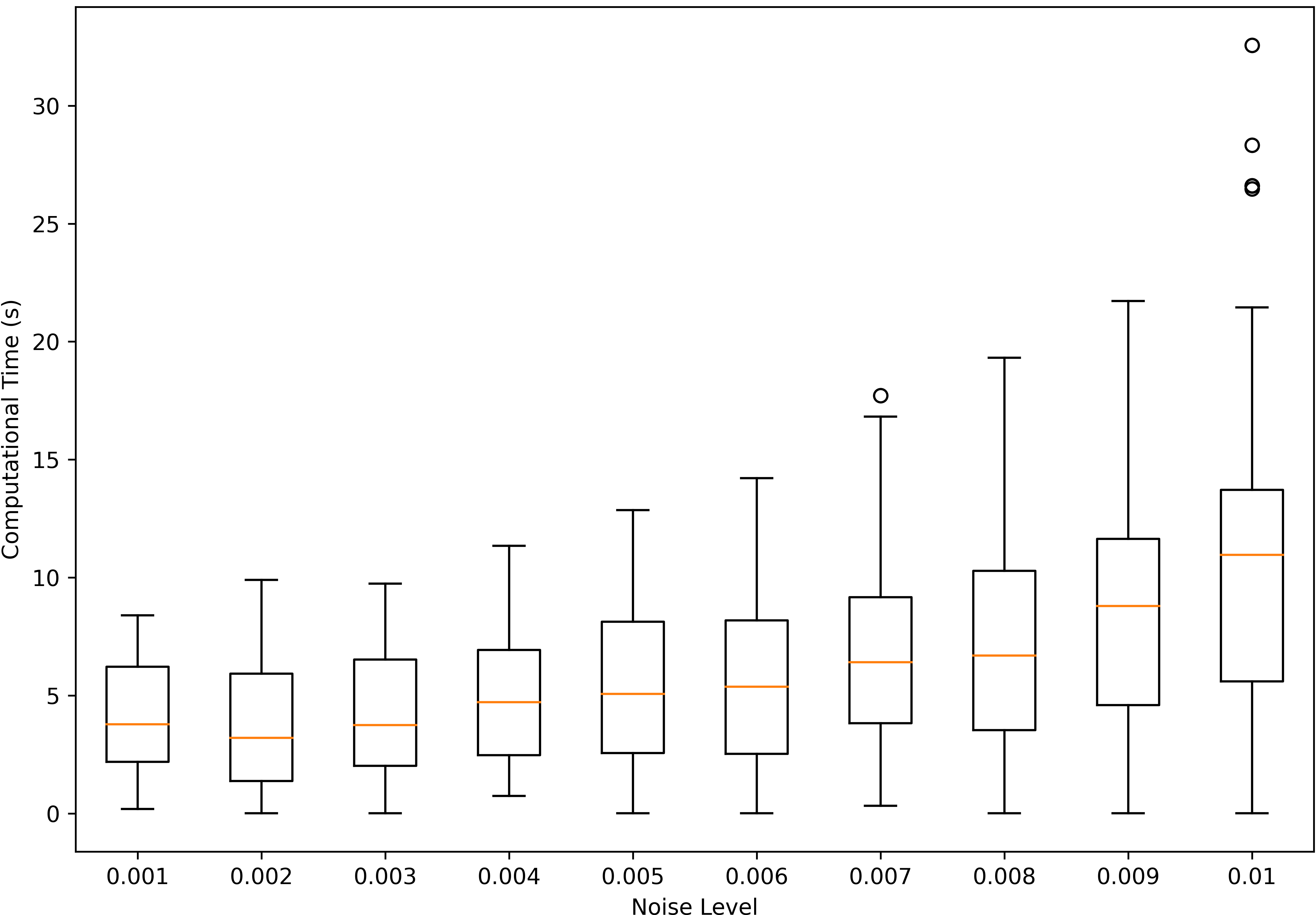}
        \caption{Box plot of the computational time under different noise levels.}
    \end{subfigure}
  \caption{Performance of the proposed method under Monte-Carlo simulation. The upper and lower bounds of the boxes depict the $75\%$ and $25\%$ quantiles, respectively. The top and bottom horizontal lines represent the $95\%$ and $5\%$ quantiles, respectively. The scattered dots illustrate the outliers.}
  \label{fig:sim}
\end{figure*}

The proposed algorithm is evaluated in a simulation dataset generated from the Erd{\H{o}}s-R{\'e}nyi model \cite{erdHos1960evolution}. Specifically, the full graph is generated as an Erd{\H{o}}s-R{\'e}nyi graph with $100$ nodes. Edges are connected between each pair of nodes with a probability of $0.1$ to simulate graphs in real applications where edges are sparsely presented. Each edge in the full graph is assigned with a random weight sampled from the uniform distribution $U(0, 1)$. The subgraph is consisted of $20$ nodes randomly sampled from the full graph. To ensure some edge connectivity in the subgraph, neighboring nodes whose pair-wise distance is smaller than $0.5$ are included. Each edge in the subgraph is then assigned with a normally distributed noise with zero mean and standard deviation of $\sigma$. The algorithm performance will be tested under different 
noise level $\sigma$.

A Monte-Carlo simulation study of $100$ iterations is conducted for $\sigma\in\left\{0.001, 0.002, \dots, 0.01\right\}$. In each iteration, both a full graph and a subgraph is generated according to the procedure described above. The standard deviation $\sigma$ is presumed to be known, and $\alpha$ is set as $0.025$. The proposed algorithm takes the full graph and subgraph as the input, and returns the node correspondence. When implementing the proposed method, we set $p=2$ to focus the topology matching on triangles. This is because a smaller $p$ provides a small searching space for all the feasible $p-$simplex and hence improves the algorithm efficiency. A discussion on the trade-off between the robustness of the method and the algorithm efficiency can be found in Section~\ref{sec:conc}.

The resultant node correspondence is compared with the true node correspondence, and the percentage of the correctly matched nodes is reported to evaluate the algorithm accuracy. The computational time in each iteration is also recorded to evaluate the algorithm efficiency. The distributions of the accuracy and computational time under the Monte-Carlo simulation are summarized in Figure~\ref{fig:sim}. We observe an decreasing trend of the matching accuracy as the noise level increases as expected. The averaged accuracy is above $95\%$ when the signal-to-noise ratio is above $20$ (corresponding to $\sigma=0.05$). Regarding the computational time, it takes $4$ seconds on average to seek a node correspondence between a full graph of $100$ nodes and a subgraph of $20$ nodes, implying the relatively high efficiency of the proposed method. 

We compared the proposed algorithm with nine different subgraph matching algorithms, including GraphQL \cite{he2008graphs}, CECI \cite{bhattarai2019ceci}, Neural \cite{liu2020neural}, QuickSI \cite{shang2008taming}), VF2++ \cite{juttner2018vf2++}, the interior point optimization solution of Eq(\ref{eq:quadratic}), and the continuous approximated solution of Eq(\ref{eq:quadratic}). Consequently, GraphQL\cite{he2008graphs}, CECI \cite{bhattarai2019ceci}, Neural \cite{liu2020neural} and QuickSI \cite{shang2008taming}) returned no result due to the lack of labels. VF2++ \cite{juttner2018vf2++} and the interior point optimization solver returned ``NaN''s due to the addition of noise. The continuous approximation solutions of Eq(\ref{eq:quadratic}) were computed via Gurobi. Specifically, it spent $522.94$ and $95942.45$ seconds for subgraph of node sizes $50$ and $100$, respectively. The solvers achieved a $0$ matching accuracy in both settings. All these comparisons demonstrate the advantage of the proposed method.

\begin{figure*}[tp]
  \centering
    \begin{subfigure}[t]{1.0\textwidth}
        \centering
        \includegraphics[height=3in]{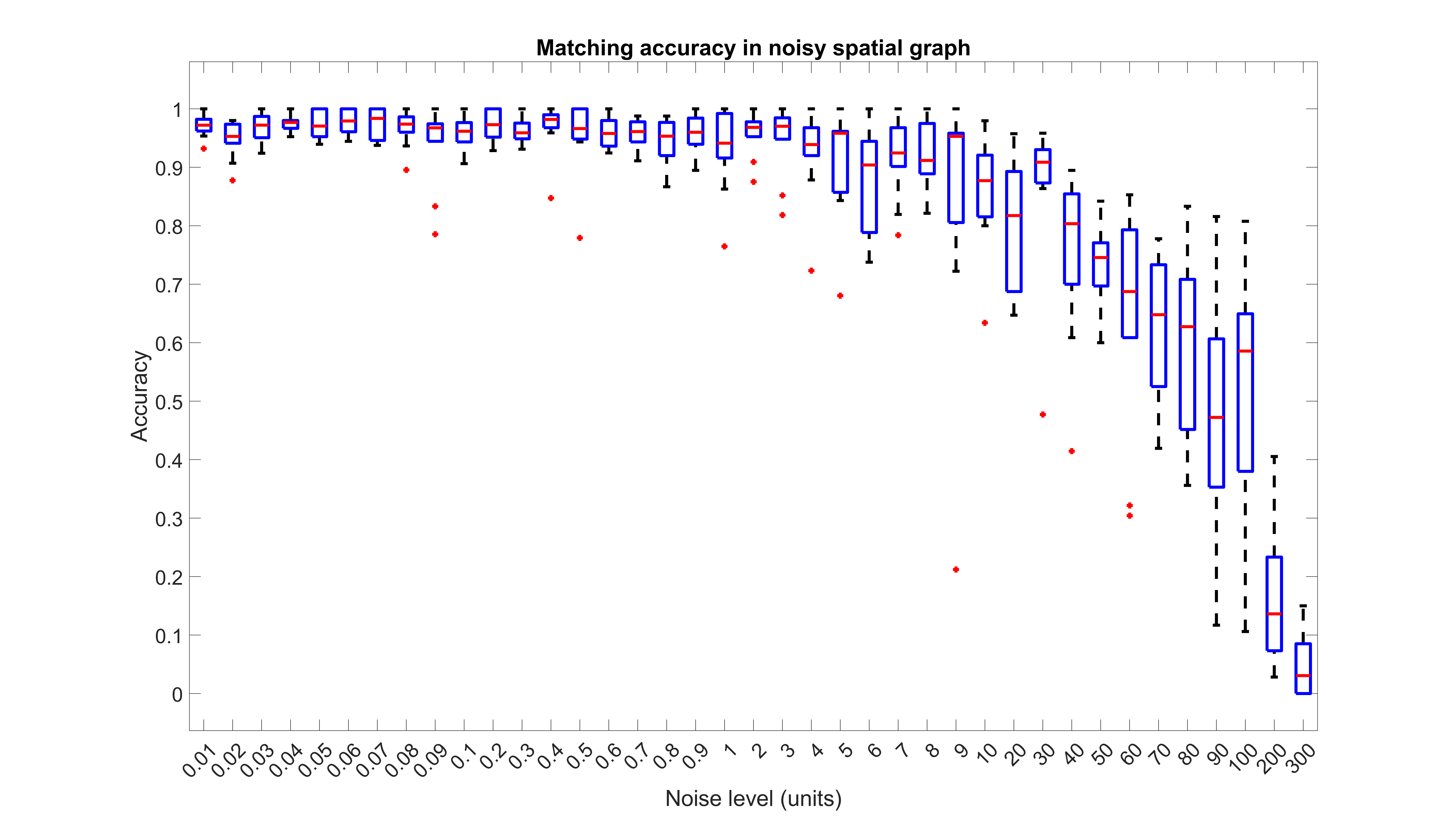}
        \caption{Box plot of the matching accuracy under different noise levels.}
    \end{subfigure}
    \begin{subfigure}[t]{1.0\textwidth}
        \centering
        \includegraphics[height=3in]{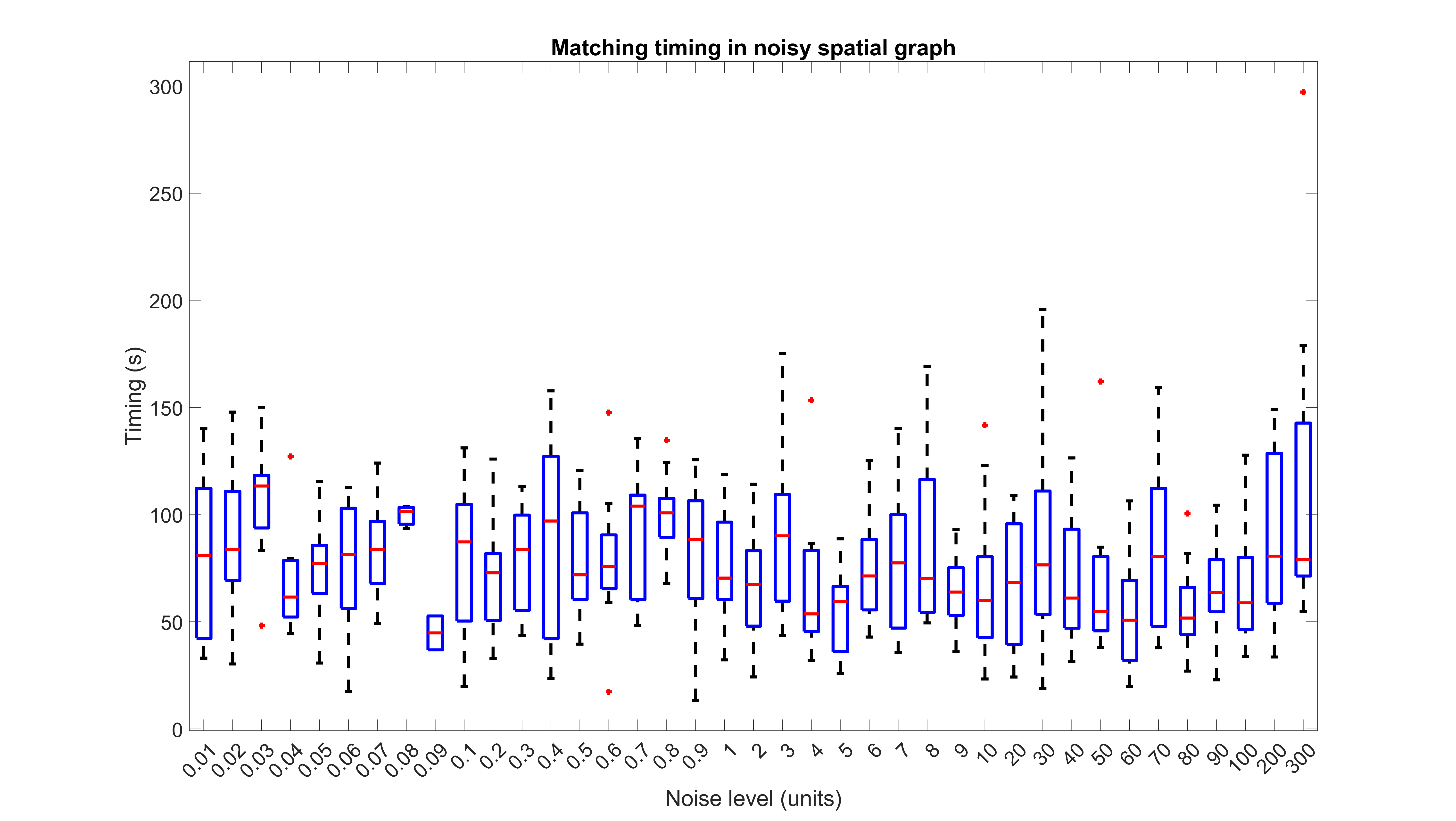}
        \caption{Box plot of the computational time under different noise levels.}
    \end{subfigure}
  \caption{Performance of the proposed method for matching spatial graphs. The upper and lower bounds of the boxes depict the $75\%$ and $25\%$ quantiles, respectively. The top and bottom horizontal lines represent the $95\%$ and $5\%$ quantiles, respectively. The scattered dots illustrate the outliers.}
  \label{fig:spatial_sim1}
\end{figure*}

\begin{figure*}
    \centering
    \includegraphics[width=1.0\textwidth]{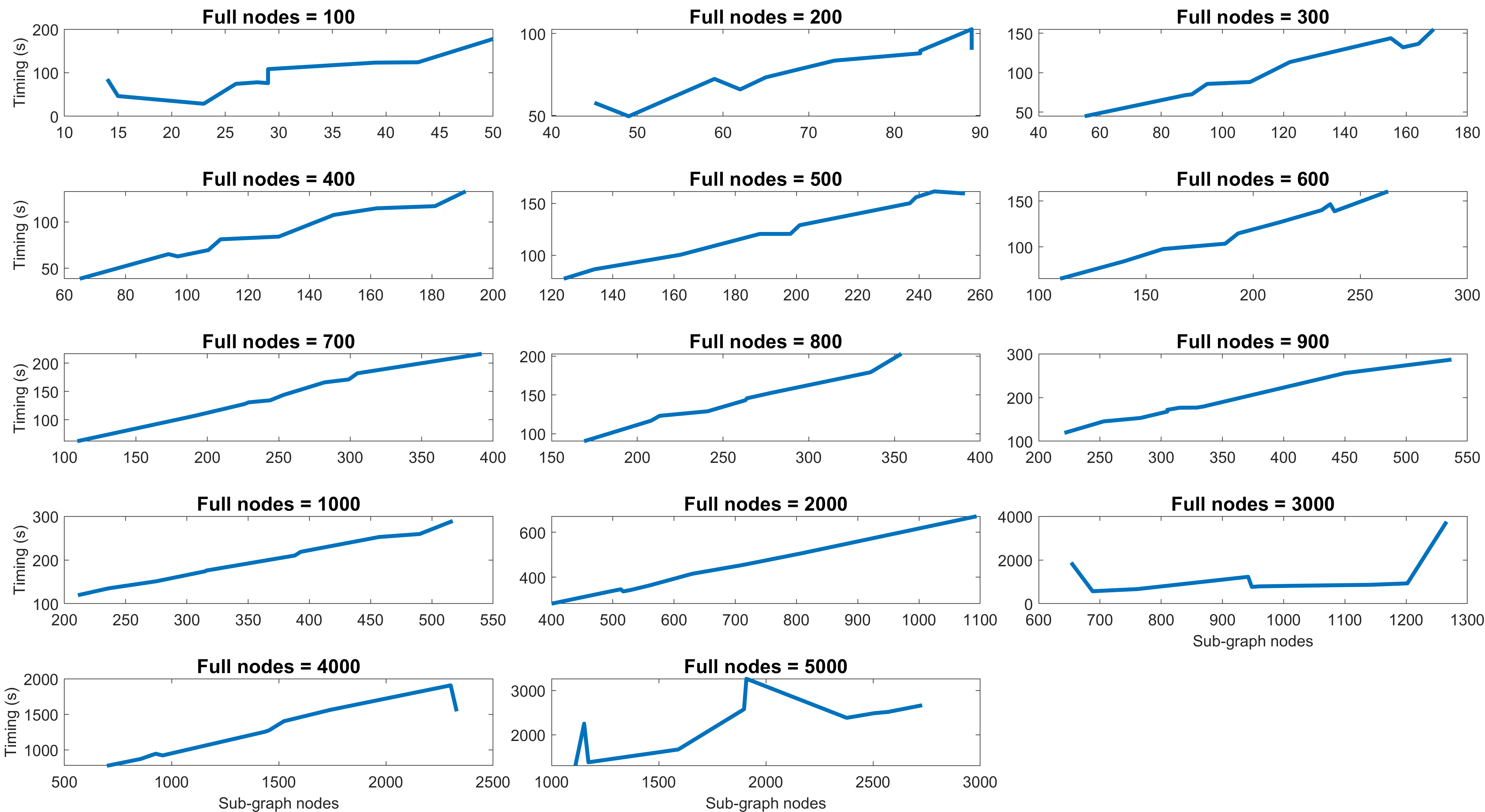}
    \caption{Timing of our proposed algorithm for different number of nodes in the full graph from 100 to 5000 with 10 runs for each node number. The number of sub-nodes chosen is based on the radius around a random node. We plot the timing with respect to the number of subgraph nodes $V_s$ chosen to show the linear performance given the number of nodes. However, we want to highlight that the change of timing over the increase in full graph nodes $V_f$ is sub-linear as can be seen from the range of the timings in the different subplots.}
    \label{fig:spatial_sim2}
\end{figure*}

\subsection{Spatial subgraph matching}
In order to further evaluate our proposed approach, we conduct an additional simulation study that involves testing on spatial graphs. By incorporating spatial graphs into our simulation, we aim to provide a more realistic and applicable setting for our approach. Spatial graphs are commonly used to represent real-world structures and networks where spatial relationships and geographic information play a crucial role.

In the initial experiment, we draw 10,000 points from a two-dimensional uniform distribution, using a scale of 10,000 units. We then construct the full graph by implementing a k-nearest neighbors graph, where each node has five neighbors. We introduce perturbations to the points by adding random noise with standard deviations ranging from {0.01, 0.02,...1, 2,...10, 20,...100, 200, 300} units. A random node is selected, and we generate a subgraph within a 10,000-unit radius. As in the previous section, we assess the algorithm's accuracy by comparing node correspondences and calculating the percentage of correctly matched nodes, conducting 10 trials for each noise level, as depicted in Fig.~\ref{fig:spatial_sim1}. Additionally, we measure the computational time for each iteration to assess efficiency. Our findings reveal that the algorithm can withstand substantial noise, though performance starts to decline gradually after a noise level of 10 units and drops around 50\% accuracy at a standard deviation of 100 units.

For the second experiment, we vary the number of nodes in the full graph starting from 100 to 1000 with the step size of 100 and then to 5000 with step size of 1000. Similar to the first experiment, we select the number of nodes in the subgraph within the 10,000 unit radius and then expand as the number of nodes increase. Fig.~\ref{fig:spatial_sim2} showcases the timing ranges for different number of nodes in the full-graph $V_f$. We run the subgraph matching over each full-graph node number for 10 runs and plot the number of subgraph nodes to the computational times. This plot shows two different important results of our proposed approach - (a) the increase in the number of subgraph nodes $V_s$ corresponds to a linear increase in timing - although the rate of increase is lesser than 1. (b) the increase in timing with respect to the full graph nodes $V_f$ increases at a much lower rate as given by the ranges of the timings in the different sub-plots. This provides some empirical support for our assertions. 

\section{Template matching}
\label{sec:case}
\begin{figure*}[!h]
  \centering
  \includegraphics[width=1.0\textwidth]{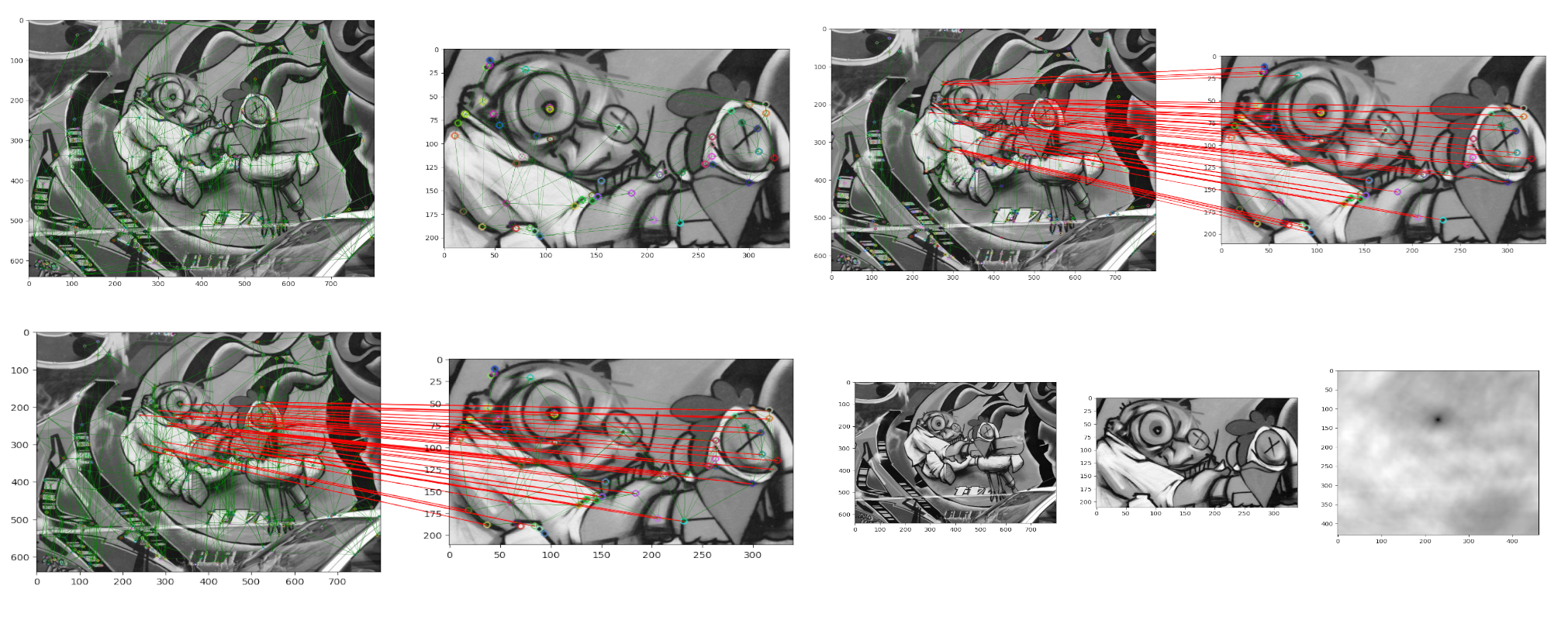}
  \caption{Top left - The sampled graffiti image in affine covariant features dataset and cropped image. The green lines depicts the constrained Delaunay triangulations of SIFT features. Top right -  The true image correspondence between the two images. The red lines highlight the node correspondence. Bottom left - The estimated image correspondences from the proposed method. Bottom right - Failed template matching using OpenCV package for cropped image - result shown in third frame.}
  \label{fig:case}
\end{figure*}
\begin{figure*}
    \centering
    \includegraphics[width=1.0\textwidth]{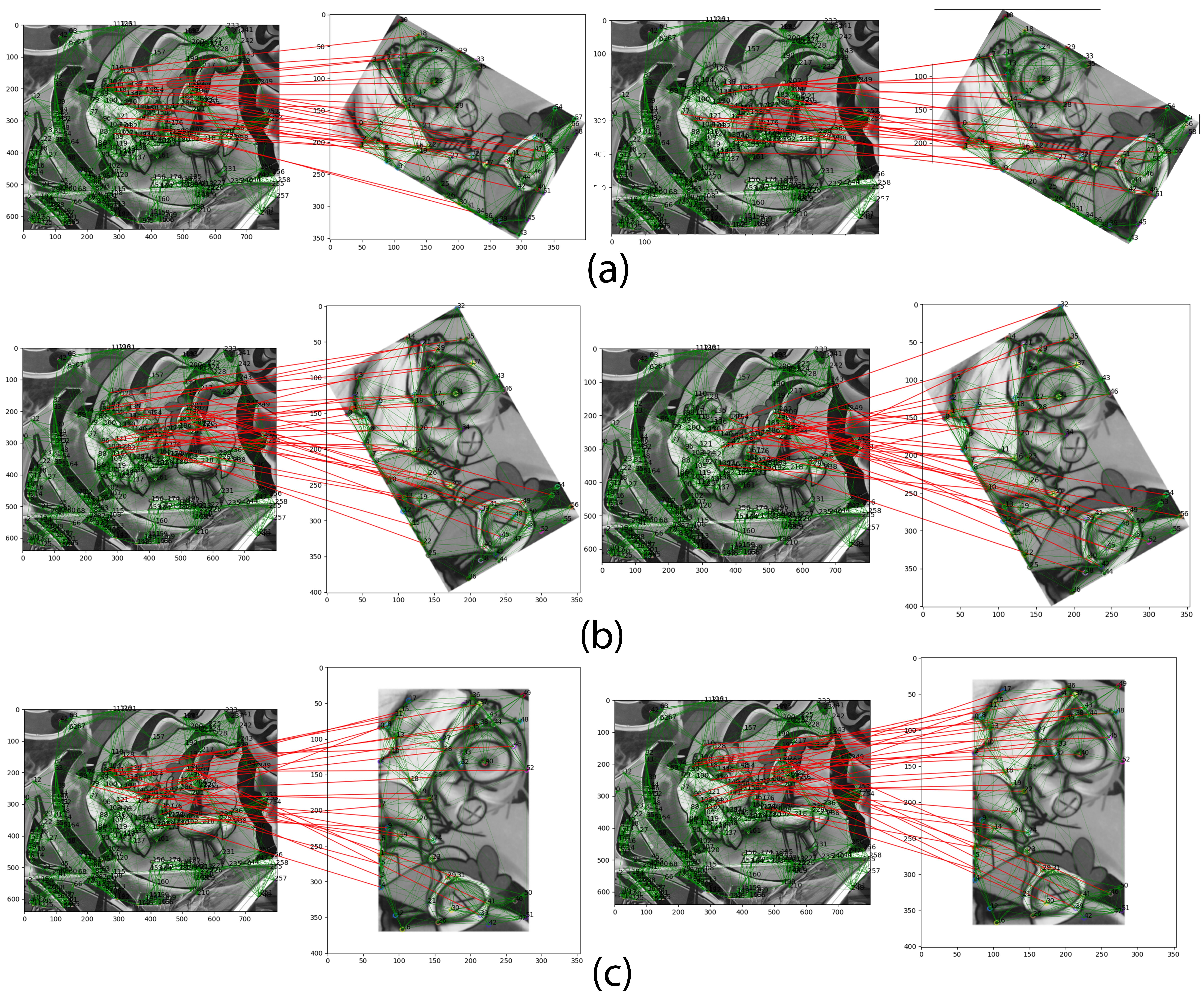}
    \caption{Matching of full image to cropped image under rotations of: (a) 30$\degree$ (b) 60$\degree$ and (c) 90$\degree$ with the true matches on left and results of our approach on the right}
    \label{fig:image_rotation}
\end{figure*}

We investigate the performance of the proposed algorithm for image correspondence using sampled images from the affine covariant features dataset \cite{mikolajczyk2005performance}. This dataset has been used as a traditional dataset for local feature detector evaluation e.g. Scale Invariant Feature Transform (SIFT) \cite{lowe1999object}, Gradient location-orientation histogram (GLOH) \cite{mikolajczyk2005performance} etc. Image correspondence or homography is used to infer the relationship between features of two or more images of the same planar object. As the first example, we create a subset of a given image by cropping part of the original image as shown in Fig \ref{fig:case}. Emulating other image graph extraction literature, we extracted keypoints from the images using the SIFT feature extraction. SIFT is known to be robust to changes in image scale, noise and illumination and thus can provide nodes which have same relative positions for the same object irrespective of changes in camera orientations. We also tried pre-trained learning based detectors but found them generally to not be effective in providing keypoints on our images. We utilize constrained Delaunay triangulations to connect nodes in order to form the graphs. The edge weights are given by the Euclidean distances between the nodes. Constraining the triangulations removes spurious triangles which increase computational load without adding new information. We show the graphs superimposed on the original and cropped image in top left sub-figure of Fig. \ref{fig:case}. Top right sub-figure of Fig. \ref{fig:case} shows the true matches between the nodes in the original and cropped image. We estimate the node correspondence by getting the coordinates of the keypoints in the cropped image and then offsetting them by the cropped origin coordinates. We then search for keypoints in the original image which lie within certain bounds and are not matched by any other keypoints. 

Bottom left of Fig. \ref{fig:case} gives the estimated correspondences using the proposed method. We choose a small set of triangles randomly in the subgraph which have paths connecting them. Based on the matching performance in the initial matching step, we tune the hyperparameter $\sigma$ and find that $\sigma=1$ pixel results in the optimal matching performance. The feasible sets are then constructed based on the constraints with the tuned $\sigma$. Given there exists a feasible set, we move to the consensus step to sequentially find nodes which are connected to the matched nodes and can be uniquely matched to nodes in the full graph. From a visual inspection, we see that even in presence of errors in SIFT keypoints, our method can robustly match nodes in a topologically consistent manner. There are couple of node matches at the top of the images which are missed due to the errors being outside our bounds.  In order to perform comparison with established algorithms, we choose OpenCV library \cite{opencv_library} which has a template matching module. The template matching module computes the match between the pixels of the template and the pixels of the image with the template being slid over the image and the loss calculated. Bottom right sub-figure of Fig. \ref{fig:case} shows the failed result with OpenCV. 

In order to showcase the global solution found by our algorithm, we present the results from rotating the cropped image by $30\degree$, $60\degree$ and $90\degree$ clockwise respectively and estimate the matching (Fig. \ref{fig:image_rotation}). Tab. \ref{tab:template} provides the precision-recall values of the template matching on the cropped and rotated images respectively. Given the rotational invariance in weights of the $2$-simplex, we see that the proposed method can match the nodes even in these kinds of large affine transformation. In all of these cases, OpenCV fails to provide any answer.

\begin{table}[h!]
\centering
\begin{tabular}{c c c}
    \hline
    Images & Precision & Recall \\
    \hline
    Cropped  & 0.91 & 0.97 \\
    Rotated-30 deg & 0.8 & 0.75 \\
    Rotated-60 deg & 0.85 & 0.7 \\
    Rotated-90 deg & 0.81 & 0.72 \\
    \hline
\end{tabular}
\caption{Precision-recall values for different affine transformations of the dataset}
\label{tab:template}
\end{table}

\section{Stereo matching}
\label{sec:stereo}

\begin{figure*}[!h]
  \centering
  \includegraphics[width=1.0\textwidth]{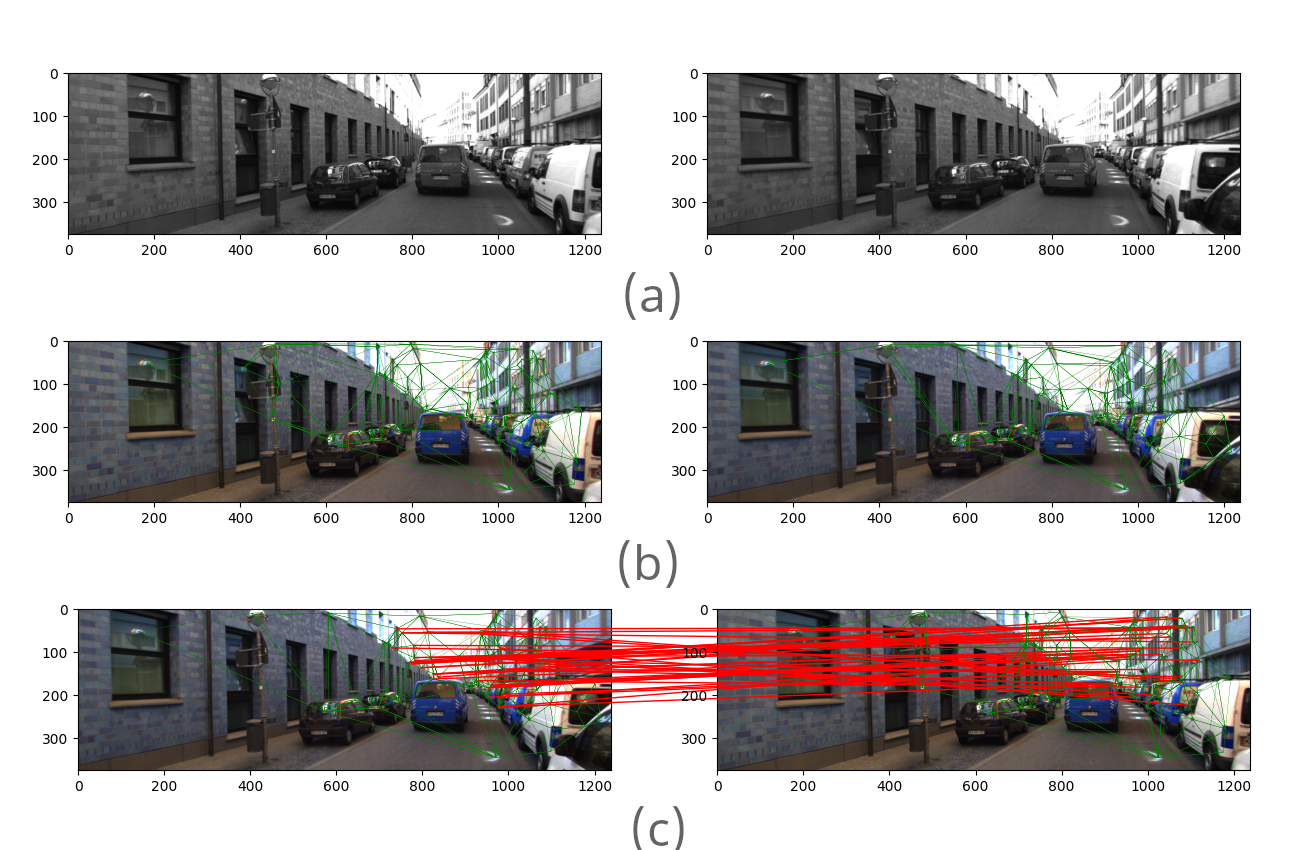}
  \caption{(a) Stereo matching of left and right images in KITTI dataset, (b) Delaunay triangulation of the SIFT features and (c) correspondences found using the proposed algorithm}
  \label{fig:stereo}
\end{figure*}
For two cameras taking images of the same scene, stereo matching is the process of finding the homologous primitives between the image pairs. In this case, we utilize the stereo-image pairs from KITTI dataset \cite{geiger2012we} from Camera 2 and 3. KITTI dataset is the first benchmark dataset for computer vision challenges in autonomous driving and has been used consistently for benchmarking algorithms in the last decade. 
 Figure \ref{fig:stereo} shows a pair of corresponding images, Delaunay triangulations on the SIFT features and finally, correspondences generated by the proposed algorithm. Visually, we can see that our algorithm can generate accurate correspondence matches while using SIFT features which tend not to be robust. We perform a quantitative analysis on the stereo matching of image pairs by estimating the transformation matrix and then find the mean rotation and translation errors. In order to perform a comparison with other approaches, we provide the estimates of Unified Sampling Consensus (USAC) \cite{raguram2012usac} in the Table \ref{tab:stereo} below:

\begin{table}[h!]
\centering
\begin{tabular}{c c c}
    \hline
    Matching algorithm & Rotation error & Translation error \\
    & [deg] & [m] \\
    \hline
    Proposed approach & 0.43 & 0.95 \\
    USAC & 0.72 & 1.2 \\
    \hline
\end{tabular}
\caption{Rotation-translation errors for the matching algorithms}
\label{tab:stereo}
\end{table}

We show that our proposed approach is able to recover the estimates of matches robustly as compared to other approaches.
\section{Conclusions}
\label{sec:conc}
We have presented a subgraph matching algorithm, providing theoretical guarantees about the accuracy of the method while looking at matching graph topology as a way to reduce the search space. Through experiments, both in simulation and real image datasets, we have presented the robustness and accuracy of the algorithm while showcasing the efficiency in simulation. The algorithm can deal with large graphs without any inherent assumptions other than weak assumptions about connectivity of the graph and presence of adequate number of simplexes. A distinctive feature of our algorithm is the concept of feasibility where an edge match is informed by the statistical threshold value found as a function of the path that the edge is part of. This can lead to substantial efficiency gains. A second distinctive feature is in the choice of the p-simplexes with the shortest path in between them as the minimum topology-preserving unit which again reduces the search space substantially. A combination of these two features leads to an algorithm that can solve an inexact subgraph matching problem with large outliers and/or deformation in sub-linear time (realistically), a large improvement over previous algorithms which are polynomial complexity, which restricts these algorithms from matching large scale graphs. 

However, our approach suffers from limitations which plague all randomized algorithms. The initial p-simplex chosen in sub-graph might not be present in the full graph due to a missing edge or might be deformed beyond the thresholding level set which can lead to wrongful initial matches. Our future work would be in understanding how to utilize topology to further remove inaccurate matches - particularly, moving away from the current technique of assuming that the initial match is the truth and constantly checking the previous matches according to global topology. Furthermore, there is a known trade-off between the algorithm's robustness and efficiency under different choices of $p$. Specifically, A large $p$ improves the algorithm robustness by comparing more edges in each topology unit. However, it sacrifices the algorithm efficiency by allowing for more feasible $p-$simplex. In this paper, the method is implemented with $p=2$ (triangles) to improve the efficiency. For more noisy graph matching problems, it is recommended to increase $p$ for better matching performances, which will be done in the future work.

%% The Appendices part is started with the command \appendix;
%% appendix sections are then done as normal sections
%% \appendix

%% \section{}
%% \label{}

%% If you have bibdatabase file and want bibtex to generate the
%% bibitems, please use
%%
%%  \bibliographystyle{elsarticle-harv} 
%%  \bibliography{<your bibdatabase>}

%% else use the following coding to input the bibitems directly in the
%% TeX file.

% \begin{thebibliography}{00}

%% \bibitem[Author(year)]{label}
%% Text of bibliographic item

% \bibitem[ ()]{}

% \end{thebibliography}
\bibliographystyle{IEEETran} 
\bibliography{main}
\end{document}